\pdfoutput=1

\documentclass[11pt]{article}

\usepackage[]{acl}
\usepackage{comment}
\usepackage{amsmath}
\usepackage{times}
\usepackage{latexsym}
\usepackage{todonotes}
\usepackage{booktabs}
\usepackage{graphicx}
\usepackage{multirow}
\usepackage[T1]{fontenc}

\usepackage[utf8]{inputenc}

\usepackage{microtype}
\usepackage{array}
\newcolumntype{x}[1]{>{\centering\let\newline\\\arraybackslash\hspace{0pt}}p{#1}}

\title{GenTUS: Simulating User Behaviour and Language\\ in Task-oriented Dialogues with Generative Transformers}

\author{Hsien-Chin Lin, Christian Geishauser, Shutong Feng, Nurul Lubis, \\ {\bf Carel van Niekerk, Michael Heck \and Milica  {Gašić}}\\
Heinrich Heine University Düsseldorf, Germany\\
  \texttt{\{linh,geishaus,shutong.feng,lubis,niekerk,heckmi,gasic\}@hhu.de}\\}
\begin{document}

\maketitle
\begin{abstract}
User simulators (USs) are commonly used to train task-oriented dialogue systems (DSs) via reinforcement learning. 
The interactions often take place on semantic level for efficiency, but there is still a gap from semantic actions to natural language, which causes a mismatch between training and deployment environment.
Incorporating a natural language generation (NLG) module with USs during training can partly deal with this problem.  However, since the policy and NLG of USs are optimised separately, these simulated user utterances may not be natural enough in a given context. 
In this work, we propose a generative transformer-based user simulator (GenTUS). GenTUS consists of an encoder-decoder structure, which means it can optimise both the user policy and natural language generation jointly. GenTUS generates both semantic actions and natural language utterances, preserving interpretability and enhancing language variation. In addition, by representing the inputs and outputs as word sequences and by using a large pre-trained language model we can achieve generalisability in feature representation.
We evaluate GenTUS with automatic metrics and human evaluation. 
Our results show that GenTUS generates more natural language and is able to transfer to an unseen ontology in a zero-shot fashion. In addition, its behaviour can be further shaped with reinforcement learning opening the door to training specialised user simulators.
\end{abstract}

\section{Introduction}

Task-oriented dialogue systems (DSs) assist their users in accomplishing a goal, such as booking a flight ticket or making a payment. This should be done through natural language interactions between the system and the user, whilst the system interacts with various external databases and API calls in the background. The core component of such a DS is the dialogue policy module, which decides what should be said to the user next. This module can be trained via interaction with users, through reinforcement learning (RL). However, this creates a conflict between the high cost of interacting with real users and the large amount of interactions required for RL. As a result, user simulators (USs) are often utilised instead to train dialogue policies, as they make it possible for the system to learn from a large number of interactions in a controlled environment at a fraction of the cost.


Rule-based USs are widely used both in research and industry because they are interpretable and can be built without a labelled dataset.
However, designing the rules demands expert knowledge and creating these rules becomes intractable on complex domains, making them only suitable for small and simple domains. In addition, human behaviour is too complex and diverse to be manually described by rules, leading to sub-optimal performance of DSs in deployment scenarios~\cite{schatzmann2006survey}.

On the other hand, data-driven USs can be built with less expert involvement. However, these models are either ontology-dependent~\cite{el2016sequence,gur2018user, kreyssig-etal-2018-neural}, which means adapting to a new domain requires re-engineering the feature representation or re-training the model, or they do not model the language of the user~\cite{lin-etal-2021-domain}.
Both shortcomings are serious. The user simulator needs to support zero-shot transfer across ontologies, as it is difficult to collect enough labelled data for each new domain. The ability to produce natural language output is also critical as it makes the training and testing environment more challenging and similar to the real user scenario. Therefore, models that can attain both properties are much needed.

In this work, we propose a model that has both desired properties. More specifically, our contributions are as follows:
\begin{itemize}
    \item We, propose a \textbf{gen}erative \textbf{t}ransformer-based \textbf{u}ser \textbf{s}imulator that we call \emph{GenTUS}\footnote{\url{https://gitlab.cs.uni-duesseldorf.de/general/dsml/gentus-public.git}}. The response of \emph{GenTUS} includes both semantic actions and natural language utterances, which retains interpretability and induces linguistic variation.
    \item By optimising the user policy and natural language jointly, GenTUS generates more natural language in the given context.
    \item GenTUS can adapt to an unseen ontology in a zero-shot fashion and have its behaviour further shaped by reinforcement learning (RL).
\end{itemize}

The rest of the paper is organised as follows. In Section~\ref{sec:related}, we review the related work. Section~\ref{sec:gentus} describes in detail the proposed simulation framework. In Section~\ref{sec:set-up}, we present the experimental set-up, followed by the experimental results in Section~\ref{sec:results}. We conclude with Section~\ref{sec:conclusion}.

\section{Related Work}\label{sec:related}

The performance of a task-oriented dialogue policy trained by RL is significantly affected by the quality of the US used to generate the interactions~\cite{schatzmann2005quantitative}. 
An N-gram user simulator proposed by \newcite{eckert1997user} is one of the earliest data-driven models. This model predicts the user action $a_u$ according to the system action $a_m$ based on a bi-gram model $P(a_u|a_m)$. 
Its behaviour is often unreasonable since it only takes the latest system action as input without any information about the user goal. 
Therefore, models which can act on a given user goal were introduced \cite{georgila2006user, eshky-etal-2012-generative}.
A Bayesian user simulation model which predicts the user action based on the user goal is proposed by \newcite{daubigney2012comprehensive}. In \newcite{cuayahuitl2005human}, the user and the system behaviour are modelled by hidden Markov models. 
A graph-based US, which constructs a graph from all possible dialogue paths, is proposed by \newcite{scheffler2002automatic}. 
This simulator can act reasonably and consistently, but it is not practical to implement in a complex scenario, as it requires extensive domain knowledge.

The agenda-based user simulator (ABUS) \cite{schatzmann-etal-2007-agenda} is widely used to train tourist-information DSs. Its behaviour is based on hand-crafted stacking and popping of rules with a stack-like agenda user goal, ordered by the priority of the user actions. It is difficult to transfer this model to a new ontology because the rules need to be redesigned. Moreover, it only provides semantic-level dialogue acts.

To reduce the involvement of experts, further data-driven user simulator approaches have been proposed. The sequence-to-sequence (Seq2Seq) model structure is the most common framework. A semantic level Seq2Seq user simulator with an encoder-decoder structure is proposed by \newcite{el2016sequence}. This model embeds the dialogue history into a context vector via a recurrent neural network (RNN) encoder. Its decoder then generates user actions based on the context embedding vector.

Instead of generating dialogue acts, the neural user simulator (NUS) of \citet{kreyssig-etal-2018-neural} can generate responses in natural language. However, this model has limited interpretability because it does not provide semantic-level outputs and its input representation is domain-dependent.

The variational hierarchical Seq2Seq user simulator (VHUS) proposed by \newcite{gur2018user} encodes the system actions and the user goal by RNNs instead of complex dialogue history features and generates semantic user actions. Its features are still domain-dependent as system actions and user goals are represented by domain-dependent one-hot encodings. As VHUS has no constraints in the decoding process, it often generates impossible actions under the given ontology.

A domain-independent transformer-based user simulator (TUS) is proposed by \newcite{lin-etal-2021-domain}. With domain-independent input and output feature representations, TUS can adapt to an unseen domain in a zero-shot fashion. However, it does not model natural language output. Moreover, all intents are part of the model, which makes transfer to an unrelated ontology, i.e.\ the one with a different sets of intents, difficult.

To convert the dialogue acts from the semantic level to natural language, a user simulator commonly includes an NLG module connected to the semantic level user policy. Although template-based NLGs are widely used in research, creating templates for every dialogue act is labour-intensive and lacks language variation. Data-driven NLG models, such as SC-LSTM \cite{wen-etal-2015-semantically} and SC-GPT \cite{peng2020few} can generate natural language utterances conditioned on given semantic actions. However, taking only semantic actions as input, their results may not be sufficiently natural in a given context. In addition, the user policy and NLG model cannot be optimised jointly within the modular architecture.

An end-to-end US which generates both dialogue acts and utterances is proposed by \newcite{tseng-etal-2021-transferable}, although in their evaluation they train a DS using only the semantic actions from the US. 
The NLG of this US is based on a simple delexicalised LSTM model.  The user goal is represented as a binary vector, with each dimension representing a domain-slot pair in the ontology. This creates several obstacles for transfer to an unseen ontology: such a transfer would require further hand-coded lexicalisation rules for the NLG component, modifications of the feature representations and further fine-tuning of the US policy.  

\section{Generative Transformer-based User Simulation}\label{sec:gentus}

\begin{figure*}[h]
    \centering
    \includegraphics[width=0.85\textwidth]{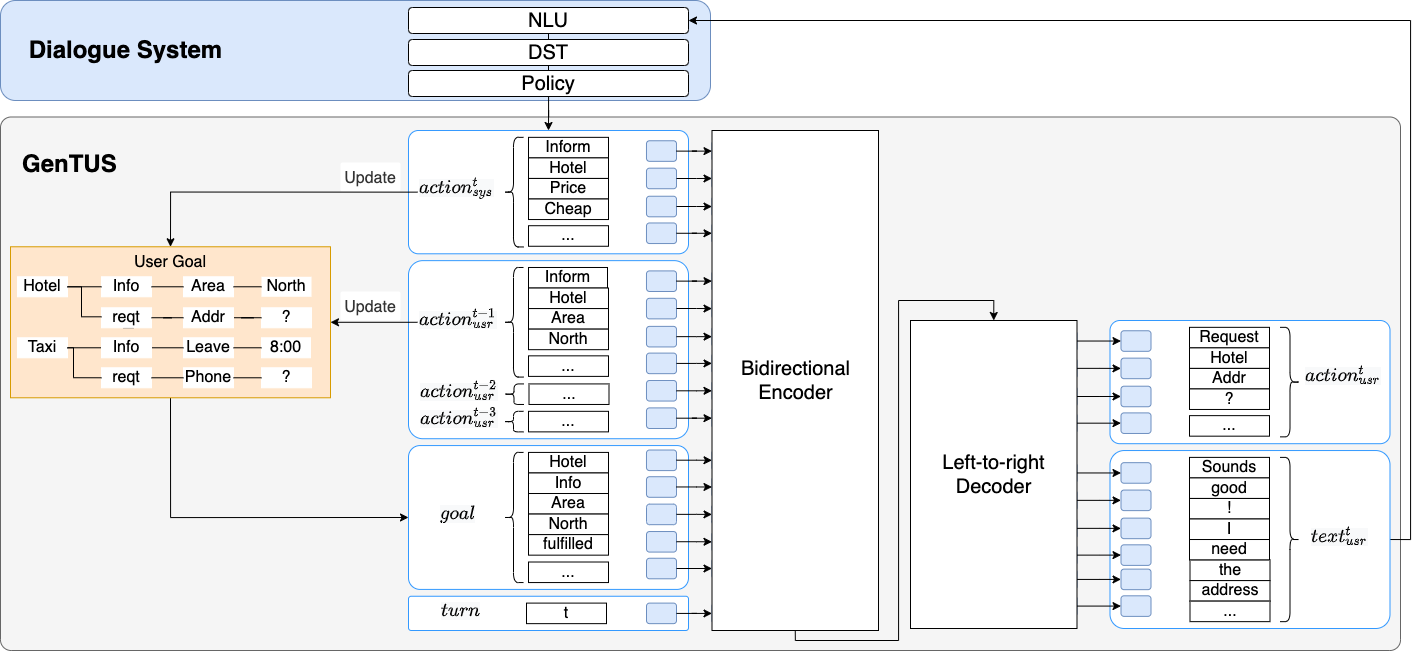}
    \caption{The model structure of GenTUS. Both input and output are JSON-formatted word sequences.}
    \label{fig:stepGenTUS}
\end{figure*}

Task-oriented DSs are expected to handle the requests of real users in natural language. Therefore, when designing USs, it is important to endow them with the ability to converse with the system via natural language as well. In this way, we can study, for example, the robustness of the systems towards misunderstandings that may occur when conversing with real users. On the other hand, users rarely misunderstand the DS response. It is hence reasonable to assume that the input to the US may be on the semantic level. This is also practical in such cases as when DSs need to execute API calls, such as playing a song or turning off the light. 

Task-oriented DSs are built upon an \emph{ontology} which includes all possible \emph{intents} that the user or the system can exhibit in their actions and \emph{domains}, which describes the entities the user or the system can talk about. Domains are further characterised by a number of \emph{slots} and each slot can take a number of \emph{values}. In task-oriented DS we assume that the user has a particular goal they want to achieve. We define goal as the following set $G = \{ domain_1: [ (slot_1, value_1), (slot_2, value_2), \dots], domain_2: [ (slot_3, value_3),  \dots], \dots\}$,
where domains, slots and values are selected from the ontology.

The semantic \emph{user action} and \emph{system action} are composed of several tuples of the following structure: $(intent, domain, slot, value)$.  Users and systems may have different intents, e.g., systems can \emph{recommend} an option and users can \emph{negate} the recommended offer. A semantic action can be converted into a natural language utterance, which we denote with $text_{usr}$ in the case of a user action.

User simulation in a task-oriented dialogue can be modelled as a sequence-to-sequence problem.
For each turn, GenTUS takes the context information as an input sequence, including the system action, the user history, the user goal, and turn information, and generates the semantic action and the natural language response as the output sequence. In following sections, we provide more details.

\subsection{Model Structure}
\label{sec:model_detail}
The backbone of the proposed GenTUS user simulation model is an encoder-decoder structure as shown in Fig.~\ref{fig:stepGenTUS}. In turn $t$, the user goal is updated by the user action from the previous turn and the current system action. If the system informs that the user's request is not possible or fails, the value of constraint slots will be replaced by a random value. The encoder takes the system action $action_{sys}^t$, user actions from previous 3 turns $action_{usr}^{t-1:t-3}$, the user goal $goal$, and the turn number $t$ as input. Then the decoder generates both the user semantic action $action_{usr}^{t}$ conditioned on the output of the encoder and the associated natural language response $text_{usr}$. We initialise GenTUS by BART~\cite{lewis2020bart}, which is a transformer-based natural language generator with a bidirectional encoder and a left-to-right decoder. BART achieves convincing results on text generation and comprehension tasks after fine-tuning. 

\subsection{Input and Output Representation}
\label{sec:feature_representation}
The \emph{system action} and \emph{user action} are semantic level dialogue acts and are represented by a list of tuples $(intent, domain, slot, value)$. Note that the output of this user simulator is a semantic as well as a natural language representation of the user action. The natural language action is sent to the system, while the semantic action is retained by the user simulator for the next turn. 
The user goal $goal$ is represented by a list of tuples,
\begin{equation}
\begin{aligned}
\relax
[&(domain_1, type_1, slot_1, value_1, status_1),\\
    &(domain_2, type_2, slot_2, value_2, status_2), \dots]
\end{aligned}
\end{equation} 
Following the setting in \newcite{lin-etal-2021-domain}, the tuples are ordered by the user preference, which means one tuple is in front of the others if the user prefer to mention it earlier. The \emph{intent}, \emph{domain}, \emph{slot}, and \emph{value} are sampled from the ontology. The \emph{type} represents whether a slot in the goal is a constraint \emph{info}, a request \emph{reqt}, or a booking information \emph{book}. The $status$ represents the condition of each domain-slot pair. It can be \emph{fulfilled}, \emph{in conflict}, \emph{requested}, or \emph{not mentioned}. The turn information is the number of the current dialogue turn. 
We represent the input to GenTUS as a JSON-formatted string: "\{"system": $action_{sys}^t$, "user": $action_{usr}^{t-1:t-3}$, "goal": $goal$, "turn": $t$\}".

The output of GenTUS is a set of semantic-level user sub-actions and the corresponding  utterance in natural language. The output is also easily represented as a JSON-formatted string: "\{"action": $action_{usr}^t$, "text": $text_{usr}^{t}$\}". 

As ultimately both input and output contain only words, we can train GenTUS as a sequence-to-sequence model. By using a pre-trained language model for initialisation, we can harness the generalisation capabilities of these powerful models when adapting to a new ontology.

\subsection{Constrained Semantic Decoding Space}
\label{sec:constrained}
\begin{figure*}[h]
    \centering
    \includegraphics[width=\textwidth]{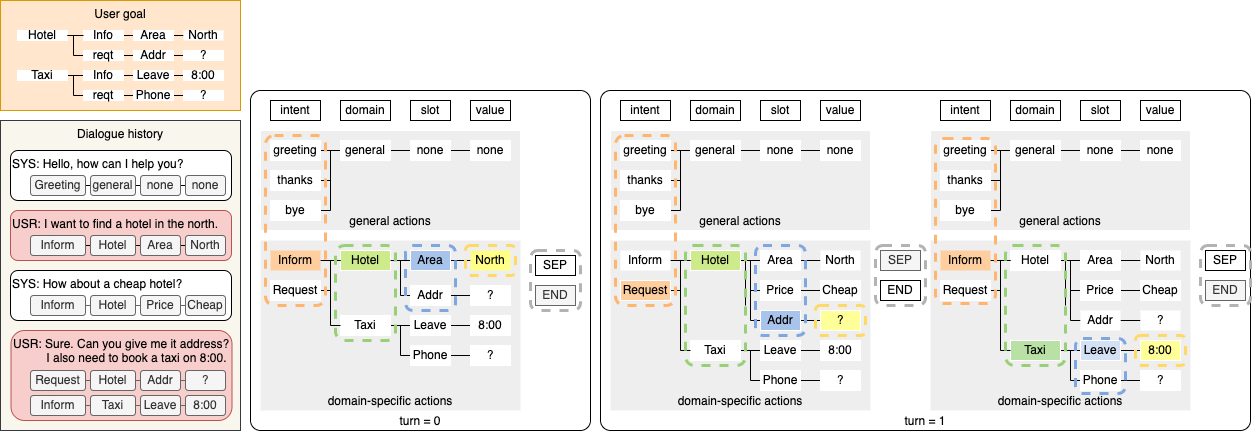}
    \caption{An example of a constrained semantic decoding space. The intents come from the ontology whereas domains, slots and values come from the user goal. In addition, system actions can insert new nodes. The user semantic actions can only contain nodes from the graph. More details are mentioned in section~\ref{sec:constrained}.}
    \label{fig:validate}
\end{figure*}

The downside of using a large pre-trained language model as a generator is that it may suffer from generating hallucinations. This means that we should place constraints on the output to prevent generating illegal semantic actions, which is particularly problematic for DSs. 

In order to only produce valid actions, every semantic action $(intent, domain, slot, value)$ is created by following a path in a graph that defines the valid actions, where the graph is constructed as follows. The possible \emph{intents} in the diagram are derived from the ontology. For example, the MultiWOZ dataset~\cite{budzianowski-etal-2018-multiwoz} contains general intents like \emph{greeting} and \emph{bye}, and domain-specific intents like \emph{inform} and \emph{request}. The possible \emph{domains}, \emph{slots} and \emph{values} are derived from the user goal, and system actions are used to update the nodes. The possible paths following intent, domain, slot and value are constrained by the ontology, which defines what valid actions are comprised of. Fig.~\ref{fig:validate} depicts an example, where GenTUS selected the action \texttt{[(Inform, Hotel, Area, North)]} in turn $0$ and \texttt{[(Request, Hotel, Addr, ?), (Inform, Taxi, Leave, 8:00)]} in turn $1$ by following the two paths in the diagram. The graph derived from the user goal is depicted on the left of Fig.~\ref{fig:validate} and updated after the system asked about a cheap hotel. After every decoded action the model can decide whether to continue or stop the decoding process. It is important to highlight that while we use the ontology to constrain the generation process, no part of the ontology is ever part of the model, but the model uses the ontology as one additional input. In that way it can be transferred to a new ontology in a purely zero-shot manner.

\section{Experimental Setup}\label{sec:set-up}

The objective of our experiments is four-fold. First, we want to show that when trained and tested on the same ontology, the user simulator can adequately capture the semantics represented in the real user data. At the same time, we also want to examine its zero-shot capability by conducting the evaluation on another unseen ontology. Second, as natural language output is an important component of the proposed model, we evaluate it separately using both automatic measures and a human preference test. Third, we jointly evaluate the GenTUS dialogue policy and its natural language output using a human trial and compare it to the state of the art. This aims to show the value of optimising the user simulator behaviour and language at the same time. Finally, we show how the behaviour of GenTUS can be further shaped by RL in interaction with a DS, with the aim of demonstrating that this model can yield a number of specialised user simulators.

\subsection{Datasets}
\label{sec:dataset}
We conduct our experiments on two corpora, the Multi-Domain Wizard-of-Oz (MultiWOZ) \citep{budzianowski-etal-2018-multiwoz} and Schema-Guided Dialogue (SGD) \citep{lee2021sgd} datasets. MultiWOZ is a human-to-human conversation dataset including around 10k dialogues, one person posing as a user and the other as an operator. In this dataset, more than one domain may be involved in one dialogue, even in the same turn. SGD consists of more than 20k dialogues between humans and a virtual assistant. The ontology of MultiWOZ includes 5 intents (3 general intents, e.g., \emph{greeting} and \emph{bye}, and 2 domain-specific intent, i.e., \emph{inform} and \emph{request}) and 7 different domains, e.g. \emph{hotel} and \emph{attraction}. On the other hand, the ontology of SGD includes 11 intents (2 general intents, i.e., \emph{thank-you} and \emph{goodbye}, and 9 domain-specific intents, e.g. \emph{inform}, \emph{request}, and \emph{confirm}) and 20 different domains, e.g., \emph{bank} and \emph{music}. More details of these two datasets are listed in Appendix~\ref{app:dataset_info}.

\subsection{Supervised Learning for GenTUS}
Our model is inherited from Huggingface's transformers \citep{wolf-etal-2020-transformers} and trained on both MultiWOZ and SGD.
To measure how well GenTUS can transfer to a new ontology, the model trained on MultiWOZ is not only tested on the MultiWOZ test set but also evaluated on the SGD test set without any further fine-tuning, and vice versa. To the best of our knowledge, no other data-driven US has been tested in such a rigorous zero-shot transfer set-up.

We evaluate NLG performance by automatic metrics, including slot error rate (SER), sacre-BLEU score \citep{post-2018-call} and self-BLEU score \citep{zhu2018texygen}, and a human preference test. SER evaluates the exact matching of semantic actions in the candidate utterance. $SER = (m+h)/N$, where $N$ is the total number of slots in semantic actions, $m$ and $h$ stand for the number of missing and hallucinated slots, respectively. 
The self-BLEU is a diversity evaluation metric. For every data point we generate a sentence. Given such a sentence, we calculate a BLEU score where the reference sentences are all other generated sentences. Then we can get the self-BLEU score by averaging all these results. The lower self-BLEU score implies the higher diversity.
We conduct the human preference test on the Amazon Mechanical Turk\footnote{\url{https://www.mturk.com/}} platform. Following the setting of \citet{peng-etal-2021-soloist}, the workers are requested to rate each utterance from 1 (bad) to 3 (good) in terms of informativeness and naturalness. \emph{Informativeness} measures whether the given utterance contains all the information specified in the semantic actions. 
\emph{Naturalness} evaluates whether the given utterance is human-like. A screenshot of this questionnaire can be found in Appendix~\ref{app:preference}.

In addition, we measure how well GenTUS can fit or transfer to a dataset using precision, recall, F1 score, as well as turn accuracy on the semantic level and sacre-BLEU on the language level. 

\subsection{Training the Dialogue System with User Simulators}
USs are designed to simulate the real-world scenario for training DSs, thus USs should respond in natural language as real users' utterances. In this section, we investigate the ability of the proposed model to train a dialogue policy by interacting on the natural language level. 

The policies of different DSs are trained by proximal policy optimization (PPO) \cite{schulman2017proximal}, a simple and stable RL algorithm, with different USs, including the agenda-based US (ABUS) with template-based NLG (ABUS-T), ABUS with SC-GPT (ABUS-S), and GenTUS which generates language. Note that we do not include NLG modules in evaluation which are based on delexicalisation, such as~\citet{tseng-etal-2021-transferable}, as their performance strongly depends on the amount of hand-coding invested in defining the delexicalisation rules. The downsides of delexicalisation already became clear in early neural network dialogue state trackers~\cite{mrksic-etal-2017-neural} and are further exacerbated in natural language generation~\cite{peng2020few}. We do however include a rule-based user simulator~\cite{schatzmann-etal-2007-agenda} with a template-based NLG, noted as ABUS-T in our experiments, as the rule-based user simulator has achieved competitive results in human evaluations~\cite{kreyssig-etal-2018-neural,lin-etal-2021-domain}. Also, TUS~\cite{lin-etal-2021-domain} did not significantly outperform ABUS in the human trial, so we exclude it from the evaluation here.
 
To deal with the user response in natural language, a natural language understanding module composed with BERT \citep{devlin-etal-2019-bert} (BERTNLU) is included and a rule-based dialogue state tracker (RuleDST) is used to track the users' states for each DS. These modules, e.g., BERTNLU, RuleDST, ABUS, a template-based NLG, and SC-GPT, are provided in the ConvLab-2 framework \cite{zhu2020convlab2}.

We train policies for $200$ epochs, each of which consists of $1000$ dialogues.
The reward function gives a reward of $80$ for a successful dialogue and $-1$ for each dialogue turn, with the maximum number of dialogue turns set to $40$. For failed dialogues, an additional penalty is set to $-40$. 
Each dialogue policy is trained on $5$ random seeds.

We apply the cross-model evaluation \cite{schatztnann2005effects} to evaluate these DSs. Different USs are used to evaluate a DS which is trained with a particular US to estimate the generalisation ability. 
We also conduct an interactive human trial. For evaluation, we select the DS policy performing best on the US it was trained on. For each DS we collected 300 dialogues. The human trial is implemented with DialCrowd \cite{lee2018dialcrowd, huynh-EtAl:2022:LREC} connected to the Amazon Mechanical Turk platform. Users are provided with randomly generated user goals based on the ontology of MultiWOZ and are required to interact with DSs in natural language. 

\subsection{Fine-tuning GenTUS with RL}

Simulators purely trained using supervised learning will learn behaviour that best fits the data and most likely will result in general behaviour. As behaviour can be very different from one user to another, it is important to be able to model different user behaviours, which will in turn result in more robust policies. To this end, we further fine-tune GenTUS using RL and shape its behaviour by deploying different reward functions. In order to achieve that, for a given user action $\{(intent_i, domain_i, slot_i, value_i)\}_{i=1}^m$, we define the turn level reward $r:=-\rho_{\mathit{eff}} + \rho_{act} \cdot m$, where $\rho_{\mathit{eff}}$ and $\rho_{act}$ are hyperparameters. In addition, as for the system reward, we give a reward of $80$ for a successful dialogue and $-40$ for a failed dialogue at the very end of the dialogue. We let GenTUS interact with the rule-based dialogue system both on semantic level and optimise its behaviour using PPO. 
We test two different reward settings that are distinguished by the turn level reward: $r_1:= -5 \cdot m$ (turn level penalty and low action reward) and $r_2:= -10 + 20 \cdot m$ (turn level penalty and high action reward). 
The corresponding average returns and trained user simulators associated with the rewards are abbreviated with $R_1, R_2$ and $\mathit{User}_1, \mathit{User}_2$ respectively. We train each model on $4$ different seeds. We then take for every seed the model with highest average return on its respective reward  and evaluate on the other reward functions to obtain a cross-reward evaluation.


\section{Experimental Results}\label{sec:results}
Our experimental results can be divided into five parts. In Section~\ref{sec:ablation}, we analysis the impact of different features with an ablation study. In Section~\ref{sec:nle}, we conduct the \emph{direct} evaluation by measuring automatic metrics (SER, sacre-BLEU, and self-BLEU) and human ratings (informativeness and naturalness) from the preference test. In Section~\ref{sec:transfer}, we focus on the generalisability of GenTUS by a zero-shot ontology transfer experiment, measured by semantic level and language level metrics on two different corpora. The \emph{indirect} evaluation is in Section~\ref{sec:indirect}. We compare DSs trained by different USs with cross-model evaluation. The result from the interactive human trial is discussed in Section~\ref{sec:interactive}. In Section~\ref{sec:user_rl}, we show that it is possible to further configure the behaviour of GenTUS via RL.

\subsection{Impact of different features}
\label{sec:ablation}
We conduct an ablation study to investigate the usefulness of our proposed feature representation. The result is shown in Table~\ref{tab:ablation}. First, we measure the performance of the model which takes the turn information, the system action $action^t_{sys}$ and the user action $action^{t-1}_{usr}$ from previous turn. Without context information, the model can only achieve 0.21 turn accuracy and 0.35 F1-score. After including the user goal $goal$, the F1-score is improved by 0.30 and the turn accuracy is also improved by 0.30 absolutely. After adding more user history $action^{t-1:t-3}_{usr}$, the F1 score is also improved slightly with the same turn accuracy.

This result indicates that the context information can improve the performance especially including the user goal in the input sequence.
\begin{table}[h]
\centering
\resizebox{\columnwidth}{!}{%
\begin{tabular}{@{}lrrrr@{}}
\toprule
\textbf{Model} & \multicolumn{1}{c}{\textbf{P}} & \multicolumn{1}{c}{\textbf{R}} & \multicolumn{1}{c}{\textbf{F1}} & \multicolumn{1}{c}{\textbf{ACC}} \\ \midrule
System and user action only & 0.42 & 0.30 & 0.35 & 0.21 \\
\quad+ user goal & 0.66 & 0.64 & 0.65 & 0.51 \\
\quad\quad+ history & 0.68 & 0.66 & 0.66 & 0.51 \\ \bottomrule
\end{tabular}%
}
\caption{The GenTUS ablation experiments on MultiWOZ. We analyse the impact of different input features by measuring precision (P), recall (R), F1 score (F1), and turn accuracy (ACC).}
\label{tab:ablation}
\end{table}

\subsection{Natural Language Evaluation}
\label{sec:nle}

\begin{table}[h]
\centering
\resizebox{\columnwidth}{!}{%
\begin{tabular}{@{}lrrr@{}}
\toprule
\bf{Model} & \bf{SER} $\downarrow$ & \bf{sacre-BLEU} $\uparrow$ & \bf{self-BLEU} $\downarrow$ \\ \midrule
Human & $3.92\%$ & - & 0.77 \\
TemplateNLG & $1.67\%$ & 10.46 & 0.89 \\
SC-GPT & $5.33\%$ & 10.51 & \textbf{0.79} \\
GenTUS-golden & $5.73\%$ & \textbf{19.61} & 0.93 \\
GenTUS & \textbf{$3.97\%$} & - & 0.95 \\ \bottomrule
\end{tabular}%
}
\caption{The NLG performance on MultiWOZ. GenTUS-golden is generated based on the golden semantic actions and GenTUS is using its own semantic action prediction. The arrow direction means which trend is better.}
\label{tab:nlg-different-model}
\end{table}

\begin{table}[h]
\centering
\small
\begin{tabular}{@{}ccc@{}}
\toprule
\textbf{Model} & \textbf{Informativeness} & \textbf{Naturalness} \\ \midrule
SC-GPT & 2.50 & 2.45 \\
GenTUS & 2.55 & \textbf{2.58} \\ \bottomrule
\end{tabular}
\caption{Human preference test for NLG on MultiWOZ. The naturalness score is statistically significantly different ($p_v < 0.05$).}
\label{tab:nlg-human-evaluation}
\end{table}

The NLG performance of different models on MultiWOZ is shown in Table~\ref{tab:nlg-different-model}. TemplateNLG, SC-GPT, and GenTUS-golden generate natural language responses from golden semantic actions and their SER is calculated based on these golden semantic actions. On the other hand, the language of GenTUS is generated based on semantic actions predicted by itself, which means we can directly measure the agreement between the semantic action the simulator indented to produce and the final natural language content produced by the simulated user. The sacre-BLEU is calculated with golden utterances. 

Although data-driven NLG models have higher SER than template-based NLG, these models have better scores in BLEU. GenTUS-golden outperforms SC-GPT by 9.10 points in BLEU because our model not only takes semantic actions as input but also context information, e.g., the user goal. Moreover, there is no statistically significant difference in SER between SC-GPT and GenTUS-golden. The human preference test in Table~\ref{tab:nlg-human-evaluation} also shows that GenTUS is more natural than SC-GPT with similar informativeness. 
The diversity of the proposed model is the worst, which is not surprising as we didn’t include beam-search or sampling to keep the computational complexity as low as possible. An investigation of a method which balances the two we leave for future work.

Without golden dialogue acts in the input, the SER of GenTUS drops by 1.77\% absolute when GenTUS generates utterances from its prediction dialogue acts instead of from golden dialogue acts, which means the language-level and semantic-level outputs of GenTUS are in agreement. In other words, with the context information and its predicted semantic actions, GenTUS can generate more natural language and have fewer missing and redundant pieces of information.

\subsection{Zero-shot Ontology Transfer}
\label{sec:transfer}
The results of zero-shot ontology transfer are shown in Table~\ref{tab:cross-dataset}.
For the semantic level evaluation, GenTUS has higher precision, recall, F1 score and turn accuracy on MultiWOZ than SGD when training and testing on the same corpus. The reason is the ontology of SGD is more complicated than MultiWOZ, i.e., contains more intents, domains, slots and values as shown in Section~\ref{sec:dataset}.

The performance of GenTUS trained on MultiWOZ dropped by 0.39 on F1 score and 0.35 on turn accuracy when testing on SGD. On the other hand, GenTUS trained on SGD can still achieve 0.49 on F1 score and 0.34 turn accuracy when testing on MultiWOZ without fine-tuning on the unseen MultiWOZ ontology. In other words, GenTUS trained on SGD can get a comparable F1 score and turn accuracy on both known and unknown ontology. 

When testing and training on the same corpus, the BLEU score of GenTUS is 17.84 on MultiWOZ and 18.30 on SGD. However, when transferring to another corpus, the BLEU score drops because users in MultiWOZ and SGD have different vocabulary and language styles.  


\begin{table}[h]
\small
\resizebox{\columnwidth}{!}{%
\begin{tabular}{@{}ccrrrrr@{}}
\toprule
\multicolumn{1}{l}{\textbf{Train}} & \multicolumn{1}{l}{\textbf{Test}} & \multicolumn{4}{c}{\textbf{Semantic}} & \multicolumn{1}{c}{\textbf{Language}} \\
\multicolumn{1}{l}{\textbf{data}} & \multicolumn{1}{l}{\textbf{data}} & \multicolumn{1}{c}{P} & \multicolumn{1}{c}{R} & \multicolumn{1}{c}{F1} & \multicolumn{1}{c}{ACC} & sacreBLEU \\ \midrule
M & M & 0.68 & 0.66 & 0.66 & 0.51 & 17.84 \\
S & S & 0.60 & 0.58 & 0.58 & 0.47 & 18.30 \\
S & M & 0.51 & 0.51 & 0.49 & 0.34 & 2.70 \\
M & S & 0.30 & 0.26 & 0.27 & 0.16 & 1.86 \\ \bottomrule
\end{tabular}%
}
\caption{The cross-dataset evaluation of GenTUS based on two different corpora, MultiWOZ 2.1 (M) and Schema-Guided Dialogue dataset (S). The semantic actions and language responses generated by GenTUS are evaluated by semantic level metrics, i.e., precision (P), recall (R), F1 score (F1) and turn accuracy (ACC), and language level metric, i.e., sacre-BLEU. }
\label{tab:cross-dataset}
\end{table}

\subsection{Cross-model Evaluation}
\label{sec:indirect}
The results of cross-model evaluation are presented in Table~\ref{tab:cross-model}. The DS trained with GenTUS has the best performance when interacting with ABUS-T in a $15\%$ absolute improvement in success rate over its performance on GenTUS. On the other hand, although the DS trained with ABUS-T achieves $78\%$ success rate, its performance drops by $28\%$ absolute when evaluated by GenTUS. The DS trained with ABUS-S also performs best when interacting with ABUS-T, with $17\%$ absolute improvement in success rate interacting with ABUS-S. All three DSs achieve their best performance when evaluated by ABUS-T, which means ABUS is the easiest setting. This indicates that it may not be sufficient to simulate real world scenario with only a hand-crafted policy and a template-based NLG. 

On the other hand, the USs with data-driven NLG are more difficult for the DS to handle. The DS trained by ABUS-T performs better than the DS trained by ABUS-S because they learn from the same policy and SC-GPT has higher SER, making the DS hard to be fully optimised.

\begin{table}[h]
\centering
\small
\begin{tabular}{@{}lccc@{}}
\toprule
\textbf{US for} & \multicolumn{3}{c}{\textbf{US for testing}} \\
\textbf{training} & ABUS-T & ABUS-S & GenTUS \\ \midrule
ABUS-T & \textbf{0.78} & \textbf{0.63} & 0.50 \\
ABUS-S & 0.74 & 0.57 & 0.45 \\
GenTUS & 0.68 & 0.43 & \textbf{0.53} \\ \bottomrule
\end{tabular}
\caption{The success rates of policies trained on GenTUS, ABUS with template NLG (ABUS-T), and ABUS with SC-GPT (ABUS-S) when tested on various USs. Each pair is evaluated by 400 dialogues on 5 seeds, which is 2K dialogues in total.}
\label{tab:cross-model}
\end{table}


\subsection{Interactive Human Trial}
\label{sec:interactive}

\begin{table}[h]
\centering
\small
\begin{tabular}{@{}lrr@{}}
\toprule
\bf{US for training} & \bf{Success} & \bf{Overall} \\ \midrule
ABUS-T & 0.75 & 3.71 \\
ABUS-S & 0.79 & 3.83 \\
GenTUS & \textbf{0.86} & \textbf{4.08} \\ \bottomrule
\end{tabular}
\caption{The interactive human trial results include success rate and overall rating as judged by users. Each system is evaluated by 300 dialogues. The success rate and overall score of GenTUS are statistically significantly different from ABUS-S and ABUS-T ($p_v < 0.05$)}
\label{tab:interactive}
\end{table}

\begin{table*}[h]
\small
\centering
\begin{tabular}{@{}lrrrrr@{}}
\toprule
\bf{Models} & \bf{Success} & \bf{Avg Acts} & \bf{Turns} & $\bf{R_1}$ & $\bf{R_2}$ \\ \midrule
User 1 & $0.84\pm0.03$ & $1.33\pm0.03$ & $7.01\pm0.27$ & $33.5\pm3.5$ & $34.2\pm3.7$ \\
User 2 & $0.78\pm0.04$ & $1.81\pm0.04$ & $7.24\pm0.33$ & $4.3\pm6.2$ & $119.1\pm15.5$ \\
Supervised & $0.76\pm0.08$ & $1.39\pm0.04$ & $7.38\pm0.32$ & $30.9\pm8.2$ & $38.6\pm10.0$ \\ \bottomrule
\end{tabular}
\caption{Results after fine-tuning GenTUS using RL on three different reward functions. Results show mean and $95 \%$ confidence intervals.}
\label{tab:gentus-rl-eval}
\end{table*}

The result of the interactive human trial is shown in Table~\ref{tab:interactive}. 155 users were involved in this trial. The number of interactions per user varies from 1 to 48. A dialogue is rated as successful if the system fulfils the user's given goal. The overall rating ranges from 1 (very poor) to 5 (excellent).

The DS trained by GenTUS outperforms the DS trained by ABUS-T and the DS trained by ABUS-S both on success rate and overall rating, which shows that is beneficial to train a DS with a jointly optimised user policy and NLG. However, we cannot observe statistically significant differences between ABUS-T and ABUS-S on success and overall rating, which means including a data-driven NLG module with the rule-based US is not sufficient to train an optimal DS.

\subsection{Fine-tuning GenTUS with RL}
\label{sec:user_rl}

The results of RL training are depicted in Table~\ref{tab:gentus-rl-eval}.
We can observe that both users obtain the highest return on the respective reward function. The success rate of both user 1 and user 2 are higher than supervised model because of the success reward signal in RL. User 1, which tries to lower its number of actions, has a similar average number of actions compared to supervised model, suggesting that paid users from the corpus do not want to say more than is necessary to achieve a successful dialogue. User 2, which is rewarded for taking many actions in a turn, shows a much higher average number of actions compared to the other users, reflecting a different user behaviour -- a chatty user.  

\section{Conclusion}\label{sec:conclusion}

We propose a generative transformer-based user simulator (GenTUS), which achieves high interpretability and linguistic variation by generating both semantic actions and natural language utterances. Moreover, it produces generalisable feature representation by treating the inputs and outputs as word sequences and leveraging a large pre-trained language model.
Our results show that GenTUS generates more natural language than SC-GPT in a given context and it can transfer to an unseen ontology in a zero-shot fashion. We consolidate our findings by a number of automatic as well as human evaluations. In addition, the GenTUS behaviour can be further configured by RL with different reward functions, providing an opportunity to build specialised USs.
In future work, we hope to modify also the NLG of GenTUS via RL in order to model user sentiment or personality.

\section*{Acknowledgements}
This work is a part of DYMO project which has received funding from the European Research Council (ERC) provided under the Horizon 2020 research and innovation programme (Grant agreement No. STG2018 804636). N. Lubis, C. van Niekerk, M. Heck and S. Feng are funded by an Alexander von Humboldt Sofja Kovalevskaja Award endowed by the German Federal Ministry of Education and Research. Computing resources were provided by Google Cloud and HHU ZIM.

\bibliography{anthology,custom}
\bibliographystyle{acl_natbib}

\appendix
\section{Intents and domains in MultiWOZ and SGD}
\label{app:dataset_info}
\begin{table}[h]
\centering
\small
\resizebox{\columnwidth}{!}{%
\begin{tabular}{@{}lll@{}}
\toprule
type & system & user \\ \midrule
general & \begin{tabular}[c]{@{}l@{}}welcome, reqmore, \\ bye, thank, greet\end{tabular} & bye, thank, greet \\ \midrule
\begin{tabular}[c]{@{}l@{}}domain-\\ specific\end{tabular} & \begin{tabular}[c]{@{}l@{}}recommend, inform, \\ request,  select, book, \\ nobook, offerbook, \\ offerbooked, nooffer\end{tabular} & inform, request \\ \bottomrule
\end{tabular}%
}
\caption{Al intents in the MultiWOZ dataset.}
\label{tab:multiwoz_intent}
\end{table}

\begin{table}[h]
\centering
\small
\resizebox{\columnwidth}{!}{%
\begin{tabular}{@{}lll@{}}
\toprule
type & system & user \\ \midrule
general & goodbye, req\_more & thank\_you, goodbye \\ \midrule
\begin{tabular}[c]{@{}l@{}}domain-\\ specific\end{tabular} & \begin{tabular}[c]{@{}l@{}}inform, notify\_success,\\ request, notify\_failure,\\ confirm, offer\_intent,\\ offer, inform\_count\end{tabular} & \begin{tabular}[c]{@{}l@{}}inform\_intent, inform\\ negate\_intent, negate\\ affirm\_intent, affirm,\\ request\_alts, request, \\ select\end{tabular} \\ \bottomrule
\end{tabular}%
}
\caption{All intents in the SGD dataset.}
\label{tab:sgd_intent}
\end{table}
\begin{table}[h]
\small
\centering
\begin{tabular}{@{}ll@{}}
\toprule
dataset & \multicolumn{1}{c}{domains} \\ \midrule
MultiWOZ & \begin{tabular}[c]{@{}l@{}}attraction, hospital, hotel, police, restaurant, \\ taxi, train\end{tabular} \\ \midrule
SGD & \begin{tabular}[c]{@{}l@{}}alarm, banks, bus, calendar, events, flights, \\ homes, hotels, media, messaging, movies, \\ music, payment, rental\_cars, restaurants, \\ ride\_sharing, servies, train, travel, weather\end{tabular} \\ \bottomrule
\end{tabular}
\caption{All domains in MultiWOZ and SGD. The names of domains in SGD are normalized.}
\label{tab:domains}
\end{table}
All intents in the MultiWOZ dataset are listed in Table~\ref{tab:multiwoz_intent} and all intents in SGD dataset are listed in Table~\ref{tab:sgd_intent}. The domains in SGD follow the form of \texttt{<domain\_name>\_<number>} and the number is used to disambiguate services from the same domain~\cite{lee2021sgd}. We normalize them to domain name only. All domains in MultiWOZ and SGD and listed in Table~\ref{tab:domains}.

\section{A generated dialogue example}
Here we show a dialogue example of GenTUS interacting with a rule-based DS in Table~\ref{tab:dialog-example}. The system's semantic actions are converted to natural language by a template-based NLG.

We can notice that GenTUS provides a natural response in turn 5. This utterance is difficult for the natural language understanding model and the system is not capable to handle it. As a result, GenTUS mentions the departure and destination explicitly in turn 8.

\begin{table*}[h]
\centering
\begin{tabular}{@{}ll@{}}
\toprule
\multicolumn{2}{c}{User goal} \\ \midrule
\multicolumn{2}{l}{\begin{tabular}[c]{@{}l@{}}attraction: \{info: \{type: college\}, reqt: \{postcode: ?, entrance fee: ?\}\}, \\ hotel: \{info: \{area: north, star:0\}, reqt: \{parking:?\}, \\ taxi: \{arriveby:13:00\}, reqt:\{phone:?, car type:?\}\}\end{tabular}} \\ \midrule
\multicolumn{2}{c}{Dialogue} \\ \midrule
0-sys & Welcome to the dialogue system. How can I help you? \\
0-usr & I'm looking for a college to visit in town. \\
1-sys & \begin{tabular}[c]{@{}l@{}}There 's 18 college in the centre , anything in mind ? \\ What area would you like this in ?\end{tabular} \\
1-usr & No, but I would like the postcode and entrance fee. \\
2-sys & \begin{tabular}[c]{@{}l@{}}We have christ's college. Its postcode is cb23bu . \\ Would you be interested in visiting a college? \\ The entry fee is free. The postcode is cb23bu. \\ Did you need any further assistance today?\end{tabular} \\
2-usr & Yes, I'm also looking for a place to stay in the north. \\
3-sys & \begin{tabular}[c]{@{}l@{}}I have 13 moderate guesthouse that fit your criteria. \\ Want one of those? I 'd be happy to help.        \\ May I ask what price range and area of town you are looking for?\end{tabular} \\
3-usr & I would like a 0 star rating. \\
4-sys & \begin{tabular}[c]{@{}l@{}}I have 1 option for you. It's listed as cheap. It is a guesthouse. \\ Would city centre north b and b work for you?\end{tabular} \\
4-usr & Does it have free parking? \\
5-sys & \begin{tabular}[c]{@{}l@{}}City centre north b and b is available would you like to try that ? \\ There is wifi available at the hotel . The parking is free . \\ The hotel is 0 stars . It is a guesthouse .\end{tabular} \\
5-usr & \textbf{Great. I also need a taxi to commute between the two places.} \\
6-sys & \begin{tabular}[c]{@{}l@{}}Okay, I can help with that. \\ What are your departure site and destination? \\ Also, when would you like to leave or arrive by?\end{tabular} \\
6-usr & I need to arrive by 13:00. \\
7-sys & There are 18. Would you like me to recommend one for you? \\
7-usr & I 'd like the car type and contact number please. \\
8-sys & \begin{tabular}[c]{@{}l@{}}There are 18 entries found for "college" in the centre of town. \\ Would you like the first listing? How about the christ's college? \\ Glad I could be of assistance! If you need anything else, let me know.\end{tabular} \\
8-usr & \textbf{Yes, I need a taxi to get me from the college to the hotel by 13:00.} \\
\multicolumn{2}{l}{...} \\ \bottomrule
\end{tabular}
\caption{A dialogue example of GenTUS interacting with rule-based dialogue system. GenTUS responds in a natural way utilising anaphora in turn 5. After it finds out the system is not capable to understand, GenTUS provides the departure and destination explicitly in turn 8.}
\label{tab:dialog-example}
\end{table*}

\section{An example of human preference test}
The dialogue history, semantic actions, and utterances generated by two different systems are presented to workers on AMT platform. They are required to rate the informativeness and naturalness of these utterances respectively.
\label{app:preference}

\begin{figure*}[hb]
    \centering
    \includegraphics[width=\textwidth]{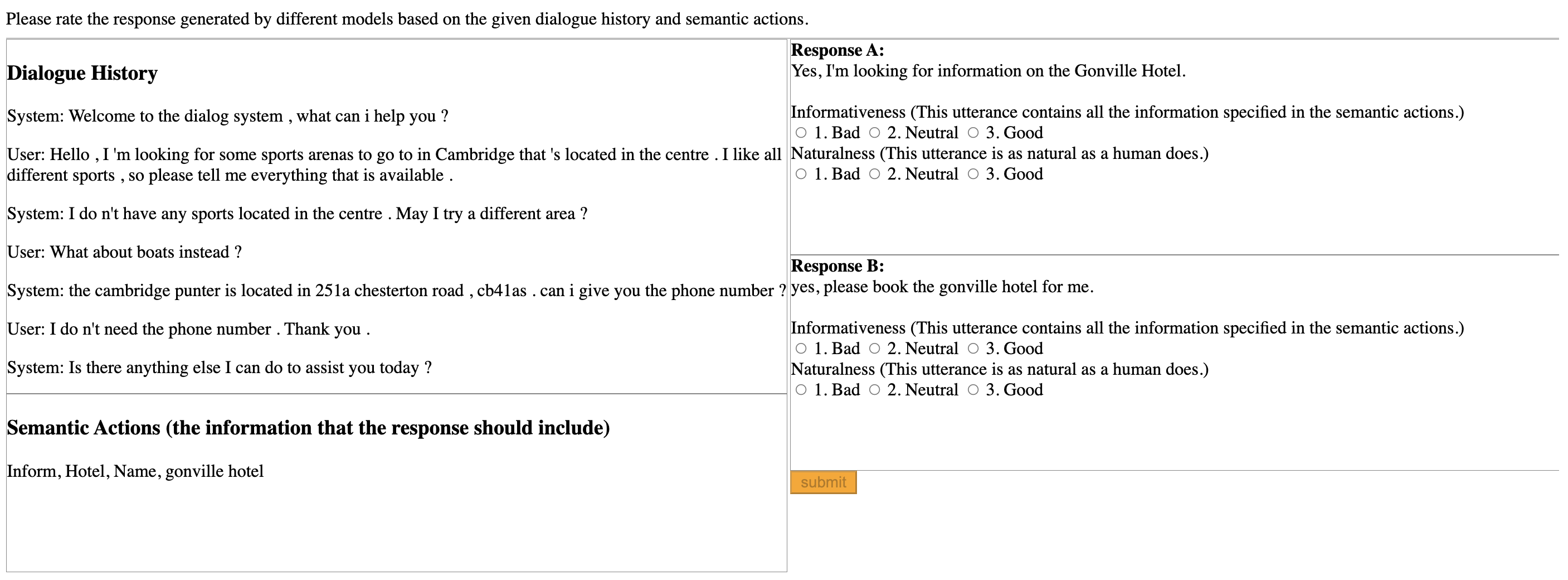}
    \caption{An example of human preference test.}
    \label{fig:preference}
\end{figure*}


\end{document}